\documentclass[11pt]{article}

\usepackage{acl}

\usepackage[T1]{fontenc}
\usepackage[utf8]{inputenc}
\usepackage{microtype}
\usepackage{inconsolata}
\usepackage{latexsym}

\usepackage{amsmath}
\usepackage{amssymb}

\usepackage{booktabs}
\usepackage{multirow}
\usepackage{tabularx}

\usepackage{graphicx}
\usepackage{pgfplots}
\pgfplotsset{compat=1.18}

\usepackage{float}
\usepackage{placeins}

\title{Understanding the Limits of Agentic ICD Coding}

\author{
  \begin{tabular}{c}
    Eng Chong Yock\textsuperscript{1,2}\quad Yushi Cao\textsuperscript{2}\quad Yiming Chen\textsuperscript{2}\\
    Kezhi Mao\textsuperscript{1}\quad Hongchao Jiang\textsuperscript{2}\thanks{Supervising author.}\\[0.3em]
    {\normalfont\normalsize\textsuperscript{1}Nanyang Technological University, Singapore}\\
    {\normalfont\normalsize\textsuperscript{2}ASUS Intelligent Cloud Services (AICS), Singapore}
  \end{tabular}
}

\begin{document}
\maketitle

% ===============================================================
\begin{abstract}
% ===============================================================

ICD-10-CM codes are alphanumeric codes used in the US to classify
diagnoses and injuries for medical billing and epidemiological
reporting.
Standard ICD-10-CM benchmarks report aggregate metrics that obscure performance on complex coding scenarios. We evaluate neural, workflow, and agentic systems on a rarity-stratified set of MIMIC-IV discharge summaries and identify two orthogonal failure modes. Neural classifiers exhibit a 0.43 micro-F1 gap between rare and common codes. Workflow systems handle rare codes well but score near zero on injury and external cause codes that require multi-step guideline following. A tool-augmented agentic configuration with structured access to official ICD-10-CM reference materials recovers up to 0.34 micro-F1 on this subset. No single system dominates across all conditions.

\end{abstract}

% ===============================================================
\section{Introduction}
% ===============================================================
Accurate ICD-10-CM coding is essential for hospital reimbursement,
epidemiological tracking, and clinical research.
The task requires assigning codes from a vocabulary of over 70,000
labels to a free-text discharge summary, subject to official sequencing
rules and chapter-specific conventions~\citep{CMS2023Guidelines}.
However, systematic analyses of failure modes in automated ICD coding remain
limited~\citep{Edin2023,Gan2025}.
 
Neural ICD coding is framed as extreme multi-label classification~\citep{Mullenbach2018,Vu2020,Huang2022}. These models
perform well on frequent codes but degrade substantially on rare
ones~\citep{Edin2023}. The ICD-10-CM label space is severely
long-tailed~\citep{PhysioNet2024}, and standard benchmarks oversample
high-frequency codes~\citep{Gan2025}, masking this degradation in
aggregate metrics.

Recent work on ICD coding~\citep{Motzfeldt2025,Akkhawatthanakun2025} has shifted toward agentic~\citep{yao2022react0,shinn2023reflexion,yang2024swe0agent0,belcak2025small} approaches that leverage large language models (LLMs) to reason, act, and use external tools during the clinical coding process. However, existing agentic systems use fixed workflows where each stage has a predetermined role. This limits performance on codes requiring multi-step guideline-following, where the order of index lookup, tabular verification, and guideline consultation depends on intermediate findings. These static architectures struggle to mimic the reasoning paths of clinical coders.

To evaluate the performance of both neural and agentic systems on ICD-10-CM coding, we curated evaluation sets of MIMIC-IV discharge summaries~\citep{PhysioNet2024} spanning codes of varying rarity and multi-step complexity. To probe whether the multi-step failure mode is structural or addressable through design, we construct a controlled agentic framework that provides an LLM with structured access to official ICD-10-CM reference materials via dedicated tools.

Our contributions are threefold: (1) We introduce a controlled multi-turn agentic evaluation framework with structured tool access to official ICD-10-CM reference materials. (2) Comprehensive evaluations identifying multi-step guideline-following failure mode orthogonal to rarity bias. (3) An in-depth analysis identifying complementary strengths and remaining gaps in current systems.\footnote{Related Work is presented in Appendix~\ref{appx:related-works}}

\FloatBarrier 

\begin{figure*}[t]
  \centering
  \includegraphics[width=0.75\textwidth]{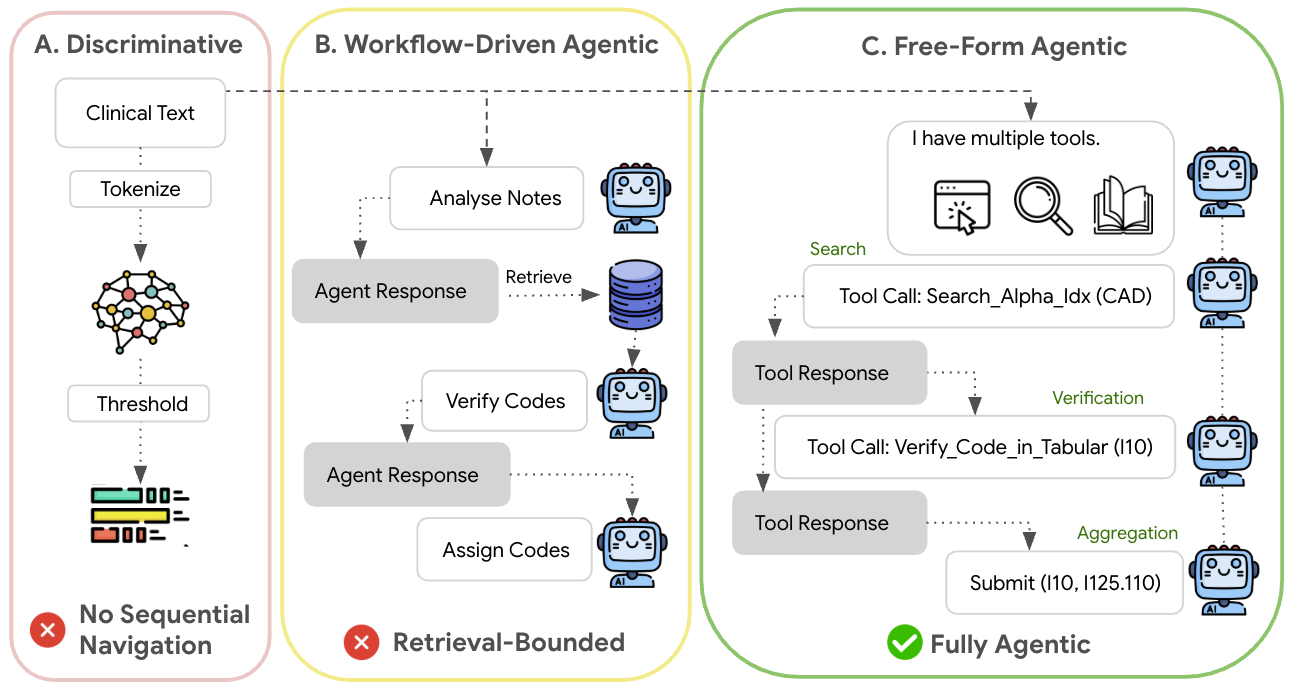}
  \caption{Comparison of three AI paradigms for automated clinical coding}
  \label{fig:agent_workflow}
\end{figure*}

% ===============================================================
\section{Evaluation Setup}
% ===============================================================

\subsection{Task Formulation}
Given a discharge summary $d$, the task is to predict a set
$\hat{Y} \subseteq \mathcal{C}$ of ICD-10-CM diagnosis codes, evaluated
against a ground-truth set $Y$ assigned by professional coders.
Correct assignment requires navigating the Alphabetical Index,
verifying codes in the Tabular List, applying chapter-specific
conventions, and restricting codes to conditions that actively affected
patient care~\citep{CMS2023Guidelines}.

\paragraph{Per-bin micro-F1.}
We compute micro-F1 independently within each rarity bin. True positives and false negatives are binned by the ground-truth code's minimum training frequency; false positives by the predicted code's. Notes can contribute to multiple bins. Confidence intervals are 95\% percentile bootstrap with 10{,}000 resamples over notes.

\subsection{Dataset and Tasks}
We evaluate on MIMIC-IV discharge summaries~\citep{johnson2023mimic4}. Uniform sampling yields evaluation sets dominated by common codes, masking performance on rare labels~\citep{Edin2023,Gan2025}, so we stratify 150 notes from the PLM-ICD test split~\citep{Huang2022} by the minimum training frequency of their codes, drawing 50 notes from each of three strata: rare (0--100), uncommon (101--1{,}000), and common ($>$1{,}000). Training frequency gives a consistent rarity axis across systems; for LLM-based systems not trained on MIMIC-IV, pre-training frequency may differ and is not measured here.

We additionally curate a multi-step navigation subset of notes with at least one ground-truth code from Chapter~19 (Injury and Poisoning; S/T codes) or Chapter~20 (External Causes; V/W/X/Y codes), yielding 153 notes and 480 instances from these chapters. Chapter~19/20 was selected as the ICD-10-CM guidelines require multiple distinct but consistent codes, making multi-step navigation an objective property of the gold annotation rather than a subjective judgment. The subset includes 117 notes spanning both chapters, 123 with a Y92 place-of-occurrence code, and 142 with a 7th-character extension.

\begin{table*}
\centering
\resizebox{0.92\textwidth}{!}{%
\begin{tabular}{l ccc ccc}
\toprule
& \multicolumn{3}{c}{\textbf{All Codes} ($n{=}150$)} & \multicolumn{3}{c}{\textbf{Multi-Step Codes} ($n{=}153$)} \\
\cmidrule(lr){2-4} \cmidrule(lr){5-7}
\textbf{Model} & \textbf{Rare} & \textbf{Unc.} & \textbf{Common} & \textbf{Rare} & \textbf{Unc.} & \textbf{Common} \\
\midrule
PLM-ICD               & .227\,{\tiny[.17,.28]} & .471\,{\tiny[.44,.51]} & \textbf{.656}\,{\tiny[.63,.68]} & .256\,{\tiny[.20,.31]} & \textbf{.375}\,{\tiny[.30,.45]} & \textbf{.435}\,{\tiny[.36,.51]} \\
\midrule
CodeSeeker (Opus 4.6) & \textbf{.560}\,{\tiny[.50,.61]} & \textbf{.483}\,{\tiny[.45,.52]} & .607\,{\tiny[.58,.63]} & .010\,{\tiny[.00,.03]} & .027\,{\tiny[.00,.07]} & .013\,{\tiny[.00,.04]} \\
CodeSeeker (GPT-5.2)  & .341\,{\tiny[.26,.42]} & .353\,{\tiny[.31,.39]} & .436\,{\tiny[.40,.47]} & .021\,{\tiny[.00,.05]} & .028\,{\tiny[.00,.07]} & .000\,{\tiny[.00,.00]} \\
\midrule
Claude Opus 4.6       & .354\,{\tiny[.30,.41]} & .425\,{\tiny[.39,.46]} & .577\,{\tiny[.55,.60]} & \textbf{.276}\,{\tiny[.22,.33]} & .363\,{\tiny[.28,.45]} & .280\,{\tiny[.21,.35]} \\
GPT-5.2               & .329\,{\tiny[.27,.39]} & .368\,{\tiny[.33,.40]} & .538\,{\tiny[.51,.57]} & .190\,{\tiny[.13,.25]} & .250\,{\tiny[.17,.33]} & .265\,{\tiny[.19,.34]} \\
GLM-5 Turbo           & .304\,{\tiny[.26,.35]} & .357\,{\tiny[.32,.39]} & .554\,{\tiny[.53,.58]} & .154\,{\tiny[.11,.20]} & .296\,{\tiny[.21,.38]} & .183\,{\tiny[.11,.25]} \\
Qwen 3.5 Plus         & .281\,{\tiny[.23,.33]} & .364\,{\tiny[.33,.40]} & .546\,{\tiny[.52,.58]} & .220\,{\tiny[.16,.29]} & .291\,{\tiny[.20,.38]} & .249\,{\tiny[.18,.32]} \\
Kimi K2.5             & .217\,{\tiny[.18,.26]} & .314\,{\tiny[.28,.35]} & .493\,{\tiny[.47,.52]} & .160\,{\tiny[.10,.22]} & .218\,{\tiny[.14,.30]} & .178\,{\tiny[.12,.24]} \\
Mistral Large 3        & .209\,{\tiny[.17,.25]} & .275\,{\tiny[.24,.31]} & .485\,{\tiny[.45,.52]} & .156\,{\tiny[.11,.20]} & .209\,{\tiny[.14,.28]} & .186\,{\tiny[.13,.25]} \\
\bottomrule
\end{tabular}}
\caption{Micro-F1 across rarity bins for all codes (150 notes, 50 per bin)
and the multi-step navigation subset (153 notes; 193/142/145 code
instances in the rare/uncommon/common bins). \textbf{Unc.}: uncommon
(101--1{,}000 training occurrences); \textbf{Common}: $>$1{,}000.
Point estimate with 95\% bootstrap CI in brackets. Best point estimate
per column in bold.}
\label{tab:results}
\end{table*}

\subsection{Evaluated ICD Systems}
We evaluate three coding systems (Figure~\ref{fig:agent_workflow}):

\paragraph{Discriminative.} PLM-ICD~\citep{Huang2022} is the current neural state of the art,
framing ICD coding as multi-label classification with label-wise
attention over a RoBERTa encoder pre-trained on biomedical text. We train the model on MIMIC-IV and evaluate it at its best-micro-F1 threshold ($\tau$=0.45).

\paragraph{Workflow-Driven Agentic.} CodeSeeker is an agentic workflow with a fixed Analyze-Locate-Assign-Verify pipeline incorporating official coding guidelines~\citep{Motzfeldt2025}. The fixed structure limits dynamic re-consultation of guidelines or the index when assigning interdependent codes.

\paragraph{Free-Form Agentic.}

To isolate the contribution of dynamic tool-use orchestration, we construct a controlled agentic configuration with structured access to official ICD-10-CM references through nine tools across four categories. \textbf{Search} covers Alphabetical Index lookup with drill navigation, the Drug, Neoplasm, and External Cause tables, and child-code browsing. \textbf{Verification} retrieves the full Tabular List entry for a candidate code, including coding rules, and accepts only codes previously seen via search. \textbf{Reference} retrieves the Official Coding Guidelines by section. \textbf{Submission} tracks all codes encountered during the session and rejects any not previously discovered, blocking hallucinated outputs. The framework enforces \emph{discovery grounding} at the tool layer: a code cannot be submitted unless returned by an earlier search or browse. Companion-code sequencing requirements are communicated via the system prompt rather than enforced by the tools.

% ===============================================================
\section{Results}
% ===============================================================

\subsection{All Codes Performance}

The left half of Table~\ref{tab:results} reports micro-F1 across label frequency bins for all codes. On rare codes, the strongest LLM-based systems outperform PLM-ICD. PLM-ICD scores 0.227 on rare codes versus 0.656 on common, consistent with training-distribution bias. CodeSeeker with Claude Opus 4.6 achieves the highest rare-code score (0.560). On common codes, PLM-ICD leads all systems.

\subsection{Multi-Step Codes Performance}
The right half of Table~\ref{tab:results} reports performance on the multi-step navigation subset. CodeSeeker performs near zero across all bins (0.010/0.027/0.013 with Claude Opus 4.6), consistent with its fixed retrieval set being unable to surface paired external-cause and place-of-occurrence codes. The agentic framework with Claude Opus 4.6 achieves the highest point estimates on rare and uncommon bins (0.276 and 0.363), though confidence intervals are wide given the subset size. PLM-ICD outperforms both despite performing no navigation, indicating that index traversal without sequential enforcement is insufficient.
% ===============================================================
\section{Discussion}
% ===============================================================

\begin{figure}[h]
\centering
\noindent\hspace*{-5mm}% 
\begin{tikzpicture}
\begin{axis}[
    width=0.95\columnwidth, height=4.4cm,
    xbar,
    bar width=4pt,
    xmin=0, xmax=0.5,
    xlabel={Median relative discovery position},
    xlabel style={font=\small},
    ytick={1,2,3,4,5,6},
    yticklabels={
    Claude\\Opus 4.6,
    GPT-5.2,
    GLM-5,
    Mistral\\Large 3,
    Qwen 3.5\\Plus,
    Kimi K2.5
    },
    yticklabel style={
    font=\tiny,
    align=right,
    text width=1.55cm
},
    yticklabel style={font=\tiny, align=right},
    y dir=reverse,
    grid=major,
    grid style={gray!20},
    legend style={font=\scriptsize, at={(0.5,1.03)}, anchor=south,
                  draw=none, legend columns=-1, column sep=1.2em},
    clip=false,
    enlarge y limits=0.08,
]
\addplot[fill=green!60!black, draw=none, bar shift=3pt] coordinates {
    (0.129,1)(0.279,2)(0.125,3)(0.279,4)(0.219,5)(0.115,6)
};
\addlegendentry{Correct (TP)}
\addplot[fill=red!70!black, draw=none, bar shift=-3pt] coordinates {
    (0.208,1)(0.385,2)(0.188,3)(0.442,4)(0.359,5)(0.211,6)
};
\addlegendentry{Wrong (FP)}
\end{axis}
\end{tikzpicture}
\vspace{-0.5em}
\caption{Distribution of median relative discovery position for correct vs.\ incorrect codes. Statistical test in Table~\ref{tab:temporal-position}.}
\label{fig:temporal-position}
\end{figure}
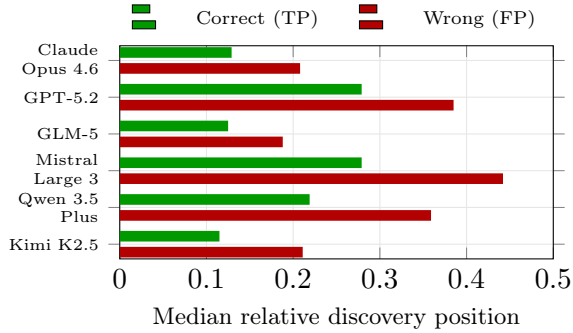

\paragraph{Correct codes are discovered early.}
Correct codes are discovered earlier than incorrect ones across models when discovery turn is normalised by session length (Figure~\ref{fig:temporal-position}). The gap is robust (Mann--Whitney $U$, $p < 10^{-13}$ for all models; Cliff's $\delta \in [0.20, 0.30]$; Appendix~\ref{app:temporal-stats}), suggesting discovery order is a reliable signal for post-hoc precision improvement.

\paragraph{Selection is harder than verification.}
Models verify 63K--178K candidate codes per run but submit less
than 4\%, of which 36--56\% are correct (Table~\ref{tab:coding-funnel}). Verification provides no filtering signal. The bottleneck is selection. More tool calls do not translate into higher precision.

\begin{table}[h]
\centering
\scriptsize
\setlength{\tabcolsep}{3pt}
\resizebox{0.92\columnwidth}{!}{%
\begin{tabular}{l r r r r}
\toprule
Model & Disc. & Sub\% & Prec\% & Calls/TP \\
\midrule
Claude Opus 4.6      &  67K & 2.8 & 56 &  8.7 \\
GPT-5.2       &  63K & 3.1 & 50 &  8.4 \\
GLM-5 Turbo   & 124K & 1.7 & 49 & 11.4 \\
Mistral Large 3 &  72K & 3.6 & 36 & 12.0 \\
Qwen 3.5 Plus & 117K & 1.7 & 49 & 10.9 \\
Kimi K2.5     & 178K & 1.6 & 36 & 13.5 \\
\bottomrule
\end{tabular}}

\caption{Coding funnel. Sub\% = submitted / discovered. Prec\% = TP /
submitted. Calls/TP = tool calls per correct code.}
\label{tab:coding-funnel}
\end{table}

\paragraph{Per-stage failure decomposition on multi-step codes.}
To localise where the framework breaks down on multi-step codes, we
trace each missed code through the four pipeline stages
(discovery $\rightarrow$ verification $\rightarrow$ submission
$\rightarrow$ specificity), broken down by chapter
(Table ~\ref{tab:multistep-failure}).

\begin{table}[h]
\centering
\scriptsize
\setlength{\tabcolsep}{3pt}
\begin{tabular}{l r r r}
\toprule
\textbf{Chapter} & \textbf{Disc.} & \textbf{Verif.} & \textbf{Sub./Spec.} \\
\midrule
Ch.\,19 S/T (injury)        & 66\% &  9\% & 3\% \\
Ch.\,20 V/W/X (mechanism)   & 86\% &  0\% & 0\% \\
Ch.\,20 Y92 (place)         & 83\% &  6\% & 0.3\% \\
Ch.\,20 Y-other             & 79\% & 14\% & 0\% \\
\bottomrule
\end{tabular}
\caption{Per-stage failure decomposition for missed multi-step codes,
by chapter.}
\label{tab:multistep-failure}
\end{table}

Three findings localise the bottleneck. \textbf{Discovery dominates}: 66--86\% of missed multi-step codes were never returned by any search or browse tool, concentrated in Chapter~20 where the External Cause Index is underutilised and Y92 place-of-occurrence codes are rarely retrieved (guideline-mandated rather than cued by the note). \textbf{Verification is not the bottleneck}: only 0--3\% of misses are wrong-7th-character errors; once the base code is retrieved, tabular verification suffices to assign the correct extension. \textbf{False positives are judgment errors, not leakage}: 74--95\% are codes the agent verified and submitted anyway, indicating the discovery gate holds and residual error is post-verification clinical judgment. The multi-step gain over CodeSeeker is attributable to tool-layer access to the External Cause Index and prompt-layer guidance on companion-code sequencing, both targeting the discovery stage.

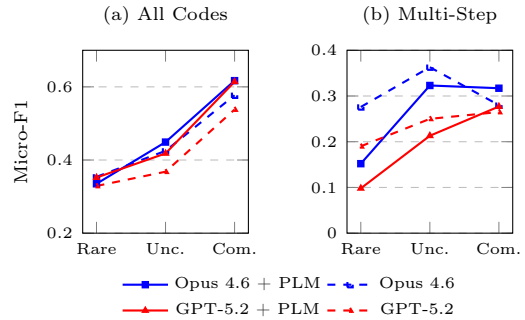
\begin{figure}[h!]
\centering
\begin{minipage}[t]{0.49\columnwidth}
\begin{tikzpicture}
\begin{axis}[
    width=\linewidth, height=4cm,
    symbolic x coords={Rare, Unc., Com.},
    xtick=data,
    ylabel={Micro-F1},
    ymin=0.2, ymax=0.7,
    ymajorgrids=true,
    grid style=dashed,
    xticklabel style={font=\tiny},
    yticklabel style={font=\tiny},
    ylabel style={font=\scriptsize},
    title={\scriptsize (a) All Codes},
    legend style={draw=none, font=\tiny, legend columns=2},
    legend to name=priminglegend,
]
\addplot[blue, mark=square*, thick, mark size=1pt] coordinates {(Rare,0.335)(Unc.,0.449)(Com.,0.617)};
\addlegendentry{Opus 4.6 + PLM}
\addplot[blue, mark=square, dashed, thick, mark size=1pt] coordinates {(Rare,0.354)(Unc.,0.425)(Com.,0.577)};
\addlegendentry{Opus 4.6}
\addplot[red, mark=triangle*, thick, mark size=1pt] coordinates {(Rare,0.352)(Unc.,0.418)(Com.,0.614)};
\addlegendentry{GPT-5.2 + PLM}
\addplot[red, mark=triangle, dashed, thick, mark size=1pt] coordinates {(Rare,0.329)(Unc.,0.368)(Com.,0.538)};
\addlegendentry{GPT-5.2}
\end{axis}
\end{tikzpicture}
\end{minipage}%
\hfill
\begin{minipage}[t]{0.49\columnwidth}
\begin{tikzpicture}
\begin{axis}[
    width=\linewidth, height=4cm,
    symbolic x coords={Rare, Unc., Com.},
    xtick=data,
    ymin=0, ymax=0.4,
    ymajorgrids=true,
    grid style=dashed,
    xticklabel style={font=\tiny},
    yticklabel style={font=\tiny},
    title={\scriptsize (b) Multi-Step},
]
\addplot[blue, mark=square*, thick, mark size=1pt] coordinates {(Rare,0.152)(Unc.,0.323)(Com.,0.317)};
\addplot[blue, mark=square, dashed, thick, mark size=1pt] coordinates {(Rare,0.276)(Unc.,0.363)(Com.,0.280)};
\addplot[red, mark=triangle*, thick, mark size=1pt] coordinates {(Rare,0.098)(Unc.,0.213)(Com.,0.277)};
\addplot[red, mark=triangle, dashed, thick, mark size=1pt] coordinates {(Rare,0.190)(Unc.,0.250)(Com.,0.265)};
\end{axis}
\end{tikzpicture}
\end{minipage}
\vspace{0.2em}
\centering\pgfplotslegendfromname{priminglegend}
\vspace{-0.5em}
\caption{Performance of PLM-ICD as a tool.}
\label{fig:priming}
\end{figure}

\paragraph{PLM-ICD as a tool.}
We investigate whether adding PLM-ICD as a tool improves agentic
performance. Using PLM-ICD hurts Claude Opus 4.6 on rare codes but
helps GPT-5.2 (Figure~\ref{fig:priming}). Claude Opus 4.6 already discovers
59\% of rare codes autonomously via 34.6 index searches per sample;
PLM-ICD reduces this by 31\%, shifting tool budget to candidate
verification instead. GPT-5.2 under-explores without PLM-ICD (49\%
rare discovery) and relies on PLM-ICD candidates to guide its search. On multi-step codes, PLM-ICD hurts both models on rare and uncommon codes and gives no gain on common codes.

% ===============================================================
% Ablation of the Agentic Framework
% ===============================================================

\section{Ablation Studies}
\label{sec:ablations}

\paragraph{Prompt-level Sequencing Instructions.}
We ablated prompt-level sequencing by removing only the mandatory external-cause sequencing instruction while holding the tool pipeline, discovery gating, budgets, and all other settings fixed. Opus 4.6 was essentially unchanged across all frequency bins, with overlapping confidence intervals and F1 differences of at most .03 (Rare: .267$\rightarrow$.241, Uncommon: .357$\rightarrow$.385, Common: .318$\rightarrow$.316). GPT-5.2 showed the same pattern on rare and uncommon codes, but common-code F1 fell sharply from .322 to .091, suggesting that explicit sequencing guidance is primarily beneficial for common cases.

\begin{table}[h]
\centering
\scriptsize
\setlength{\tabcolsep}{5pt}
\begin{tabular}{ll ccc}
\toprule
\textbf{Model} & \textbf{Config} & \textbf{Rare} & \textbf{Unc.} & \textbf{Com.} \\
\midrule
\multirow{4}{*}{Opus 4.6}
 & \multirow{2}{*}{full prompt}   & .267 & .357 & .318 \\
 &  & {\tiny[.10,.46]} & {\tiny[.21,.50]} & {\tiny[.19,.44]} \\
 & \multirow{2}{*}{seq.\ removed} & .241 & .385 & .316 \\
 &  & {\tiny[.08,.44]} & {\tiny[.24,.53]} & {\tiny[.18,.44]} \\
\midrule
\multirow{4}{*}{GPT-5.2}
 & \multirow{2}{*}{full prompt}   & .151 & .243 & .322 \\
 &  & {\tiny[.00,.30]} & {\tiny[.11,.37]} & {\tiny[.19,.44]} \\
 & \multirow{2}{*}{seq.\ removed} & .143 & .274 & .091 \\
 &  & {\tiny[.00,.31]} & {\tiny[.13,.41]} & {\tiny[.00,.20]} \\
\bottomrule
\end{tabular}
\caption{Prompt-level sequencing ablation.}
\label{tab:seq-ablation}
\end{table}

\paragraph{External Tool Access.}
We ablate the tool environment across three models, requiring the model
to assign codes from parametric knowledge alone while holding the note
set, prompt, and scoring protocol fixed
(Table~\ref{tab:tool-ablation}). Tool access helps in both settings, but
the gain falls in different places. On all codes, the largest benefit
is on rare codes, where F1 roughly doubles for every model. On multi-step codes the pattern inverts and the largest
benefit is on common codes, where Opus 4.6 improves from .050 to .280.
Recall explains the difference: without tool access, recall on common
multi-step codes falls to .021--.028 across all three models,
indicating that companion external-cause and place-of-occurrence codes
are rarely produced from memory. Precision, recall, and support are in
Appendix~\ref{app:ablation-full}.

\begin{table}[h]
\centering
\scriptsize
\setlength{\tabcolsep}{4pt}
\begin{tabular}{ll cc cc}
\toprule
& & \multicolumn{2}{c}{\textbf{All Codes}} & \multicolumn{2}{c}{\textbf{Multi-Step}} \\
\cmidrule(lr){3-4} \cmidrule(lr){5-6}
\textbf{Model} & \textbf{Band} & \textbf{Tools} & \textbf{None} & \textbf{Tools} & \textbf{None} \\
\midrule
\multirow{3}{*}{Opus 4.6}
 & Rare & .354 & .182 & .276 & .142 \\
 & Unc. & .425 & .285 & .363 & .249 \\
 & Com. & .577 & .498 & .280 & .050 \\
\midrule
\multirow{3}{*}{Qwen 3.5 Plus}
 & Rare & .281 & .153 & .220 & .110 \\
 & Unc. & .364 & .314 & .291 & .130 \\
 & Com. & .546 & .566 & .249 & .038 \\
\midrule
\multirow{3}{*}{GLM-5 Turbo}
 & Rare & .304 & .134 & .154 & .069 \\
 & Unc. & .357 & .288 & .296 & .144 \\
 & Com. & .554 & .513 & .183 & .050 \\
\bottomrule
\end{tabular}
\caption{Tool-access ablation, micro-F1. \textbf{Tools}: full nine-tool
environment. \textbf{None}: no tool access. Precision, recall, support,
and confidence intervals in Appendix~\ref{app:ablation-full}.}
\label{tab:tool-ablation}
\end{table}

\paragraph{Discovery Grounding.}
Discovery gating was not ablated because removing it changes precision
by re-admitting hallucinated codes rather than testing multi-step
navigation; instead, its role is characterised by the false-positive
analysis, where 74--95\% of false positives are verified-then-submitted,
consistent with a hallucination filter rather than the source of the
performance gains.

% ===============================================================
\section{Conclusion}
% ===============================================================
We evaluate ICD coding systems on a rarity-stratified set with
multi-step navigation cases and identify two orthogonal failure modes.
Neural classifiers degrade on rare codes due to training distribution
bias, while workflow systems score near zero on injury and external
cause codes despite strong rare-code performance. A tool-augmented agentic framework with structured access to coding guidelines and explicit prompt-level sequencing instructions partially recovers performance on this subset,  suggesting the failure is partially addressable through combining tool-layer discovery grounding and prompt-level guidance.

% ===============================================================
\section{Limitations}
% ===============================================================

Our evaluation uses MIMIC-IV discharge summaries from a single institution. ICD-10-CM code distributions vary across hospitals. MIMIC-IV labels also contain annotation noise, including coded conditions absent from the clinical note~\citep{Searle2020}, introducing uncertainty into absolute F1 values. We evaluate fixed model versions, and newer releases may change performance, particularly for agentic systems.

Our multi-step diagnostic evaluation is restricted to Chapters~19 and~20, capturing only one class of guideline-intensive coding. We therefore do not establish whether the same patterns hold across other ICD categories. The subset contains 153 notes and 480 code instances (193/142/145 across rare, uncommon, and common bins). The workflow-versus-agentic gap is large relative to this sample: CodeSeeker scores .010--.027 with Opus~4.6 versus .276--.363 for the agentic framework on the same backbone, with non-overlapping confidence intervals. Confidence intervals overlap across adjacent agentic systems, so their rankings are not statistically distinguishable.

Finally, we ablate tool access and prompt-level sequencing, but do not isolate discovery grounding or individual tools. The tool ablation
removes the entire environment rather than components such as the External
Cause Index, leaving fine-grained attribution for future work.

% ===============================================================
\section{Ethics Statement}
% ===============================================================

All clinical notes are drawn from MIMIC-IV under its established data
use agreement.
No new patient data was collected.
Model outputs are not intended for direct clinical use; all code
assignments require human expert review before deployment.

% ===============================================================
\section*{Acknowledgements}
% ===============================================================
This research was supported by the Singapore Economic Development Board (EDB) and ASUS Intelligent Cloud Services (AICS) under the Industrial Postgraduate Programme (IPP). The authors would also like to acknowledge the research facilities and academic support provided by Nanyang Technological University (NTU).

\bibliography{custom}

% ===============================================================
% ===============================================================
% ===============================================================
\appendix
% ===============================================================

% 
% ===============================================================
\section{Related Work}
\label{appx:related-works}
% ===============================================================

\paragraph{Neural classifiers.}
The dominant approach to automated ICD coding frames the task as
extreme multi-label classification~\citep{Mullenbach2018,Vu2020}.
\citet{Mullenbach2018} introduce label-wise attention over clinical
notes, establishing the foundation for subsequent work.
\citet{Huang2022} extend this with pre-trained biomedical encoders,
achieving the current neural state of the art on MIMIC-III.
\citet{Edin2023} and \citet{Gan2025} document that performance on
these benchmarks is driven by frequent codes, with rare-code
performance substantially lower. These models fail to generalise beyond the training set distribution.

\paragraph{LLM-based systems.}
Recent work has explored LLMs for ICD coding without fine-tuning
\citep{Soroush2024}, and retrieval-augmented approaches that constrain
outputs to a candidate set~\citep{Akkhawatthanakun2025}.
\citet{Motzfeldt2025} introduce CLH, an agentic workflow with a fixed Analyze-Locate-Assign-Verify pipeline that incorporates official coding guidelines. Following the established distinction between workflows (fixed orchestration of LLM calls) and agents (dynamic control flow determined at runtime), CLH falls on the workflow side: the order of analysis, location, assignment, and verification is predetermined, which constrains re-consultation of guidelines or the Alphabetical Index when assigning interdependent codes.

% ===============================================================
\section{Experimental Setup}
\label{app:setup}
% ===============================================================
\begin{table}[h]
\centering
\scriptsize
\setlength{\tabcolsep}{3pt}
\resizebox{\columnwidth}{!}{%
\begin{tabular}{lccc}
\toprule
\textbf{Model} & \textbf{Context} & \textbf{PLM} & \textbf{Reason.} \\
\midrule
\multicolumn{4}{l}{\textit{With PLM-ICD tool}} \\
\addlinespace[2pt]
Claude Opus 4.6 + PLM & 200K & $\bullet$ & $\circ$ \\
Claude Opus 4.6       & 200K & $\circ$   & $\circ$ \\
GPT-5.2 + PLM         & 1M   & $\bullet$ & $\bullet$ \\
GPT-5.2               & 1M   & $\circ$   & $\bullet$ \\
\midrule
\multicolumn{4}{l}{\textit{No PLM-ICD tool}} \\
\addlinespace[2pt]
GLM-5 Turbo           & 128K & $\circ$ & $\bullet$ \\
Mistral Large 3       & 262K & $\circ$ & $\circ$ \\
Qwen 3.5 Plus         & 131K & $\circ$ & $\circ$ \\
Kimi K2.5             & 262K & $\circ$ & $\bullet$ \\
\bottomrule
\end{tabular}}
\caption{\scriptsize Model configurations. $\bullet$/$\circ$ =
enabled/disabled. \textit{PLM}: PLM-ICD neural primer.
\textit{Reason.}: chain-of-thought reasoning. All models share
the same tool environment (200 turns, 200 tool calls, 600s timeout,
\texttt{max\_tokens}=16,384).}
\label{tab:model_configs}
\end{table}

\subsection{Hyperparameters}

All configurations share identical agentic loop parameters: maximum
200 turns, maximum 200 tool calls, and a 600-second timeout per
sample. A status message is injected every 10 turns summarising
verified codes and remaining budget. Context compression triggers
at 80\% utilisation: the system prompt and first user message are
preserved; the last 5 tool results are kept in full; older results
are compressed to one-line summaries.

\subsection{Treatment of Failed Sessions}
On timeout, model error, or stuck-loop detection, the harness
auto-submits the codes accumulated up to that point. The partial
submission is scored under the identical protocol.

\subsection{Tool Environment}
The agent has access to 9 tools in four categories
(Table~\ref{tab:tools}). The ICD-10-CM backend is built from CMS
2020 XML files: Tabular List (45,907 codes), Alphabetical Index,
Drug Table, Neoplasm Table, External Cause Index, and Official
Coding Guidelines.

\begin{table}[h]
\centering
\scriptsize
\setlength{\tabcolsep}{3pt}
\begin{tabular}{lp{3.8cm}}
\toprule
\textbf{Tool} & \textbf{Function} \\
\midrule
\multicolumn{2}{l}{\textit{Navigation (discovery-gated)}} \\
\texttt{suggest\_candidate\_codes}   & PLM-ICD Tool; disabled in no-PLM runs \\
\texttt{search\_alphabetical\_index} & Index search and hierarchical drill \\
\texttt{search\_drug\_table}         & Substance-to-code lookup \\
\texttt{search\_neoplasm\_table}     & Anatomical site-to-code lookup \\
\texttt{search\_external\_cause\_index}     & Cause, place, and activity codes \\
\texttt{browse\_code\_children}      & Subcategory drill-down \\
\midrule
\multicolumn{2}{l}{\textit{Verification (gated to discovered codes)}} \\
\texttt{verify\_code\_in\_tabular}   & Full Tabular List entry \\
\midrule
\multicolumn{2}{l}{\textit{Reference}} \\
\texttt{lookup\_coding\_guidelines}  & Guidelines by section \\
\midrule
\multicolumn{2}{l}{\textit{Submission}} \\
\texttt{submit\_codes}               & Final submission; rejects undiscovered codes \\
\bottomrule
\end{tabular}
\caption{Tool environment. Navigation tools populate the discovery
set; verification and submission are gated to that set.}
\label{tab:tools}
\end{table}

% ===============================================================
\section{Full Tool-Access Ablation}
\label{app:ablation-full}
% ===============================================================

Precision, recall, and micro-F1 for the tool-access ablation.
\textbf{Sup} denotes the number of ground-truth code instances per band; all
confidence intervals are 95\% bootstrap CIs.

\begin{table*}[t]
\centering
\scriptsize
\setlength{\tabcolsep}{4pt}
\begin{tabular}{ll r ccc ccc}
\toprule
& & & \multicolumn{3}{c}{\textbf{Full-tools}} & \multicolumn{3}{c}{\textbf{No-tools}} \\
\cmidrule(lr){4-6} \cmidrule(lr){7-9}
\textbf{Model} & \textbf{Band} & \textbf{Sup} & \textbf{Precision} & \textbf{Recall} & \textbf{F1} & \textbf{Precision} & \textbf{Recall} & \textbf{F1} \\
\midrule
\multirow{3}{*}{Opus 4.6}
 & Rare     & 233  & .317\,{\tiny[.26,.38]} & .399\,{\tiny[.33,.47]} & .354\,{\tiny[.30,.41]} & .127\,{\tiny[.10,.15]} & .322\,{\tiny[.27,.38]} & .182\,{\tiny[.15,.21]} \\
 & Uncommon & 660  & .467\,{\tiny[.43,.51]} & .391\,{\tiny[.35,.43]} & .425\,{\tiny[.39,.46]} & .245\,{\tiny[.21,.27]} & .341\,{\tiny[.30,.38]} & .285\,{\tiny[.25,.32]} \\
 & Common   & 1403 & .674\,{\tiny[.65,.70]} & .505\,{\tiny[.47,.54]} & .577\,{\tiny[.55,.60]} & .543\,{\tiny[.51,.57]} & .460\,{\tiny[.43,.49]} & .498\,{\tiny[.47,.52]} \\
\midrule
\multirow{3}{*}{Qwen 3.5 Plus}
 & Rare     & 233  & .230\,{\tiny[.19,.27]} & .360\,{\tiny[.30,.42]} & .281\,{\tiny[.23,.33]} & .107\,{\tiny[.08,.13]} & .266\,{\tiny[.21,.33]} & .153\,{\tiny[.12,.19]} \\
 & Uncommon & 660  & .389\,{\tiny[.35,.43]} & .342\,{\tiny[.31,.38]} & .364\,{\tiny[.33,.40]} & .302\,{\tiny[.27,.34]} & .326\,{\tiny[.29,.36]} & .314\,{\tiny[.28,.35]} \\
 & Common   & 1403 & .633\,{\tiny[.60,.67]} & .480\,{\tiny[.44,.52]} & .546\,{\tiny[.51,.58]} & .632\,{\tiny[.60,.66]} & .512\,{\tiny[.48,.55]} & .566\,{\tiny[.54,.59]} \\
\midrule
\multirow{3}{*}{GLM-5 Turbo}
 & Rare     & 233  & .249\,{\tiny[.20,.30]} & .391\,{\tiny[.33,.46]} & .304\,{\tiny[.26,.35]} & .099\,{\tiny[.07,.13]} & .206\,{\tiny[.15,.26]} & .134\,{\tiny[.10,.17]} \\
 & Uncommon & 660  & .362\,{\tiny[.32,.40]} & .351\,{\tiny[.31,.39]} & .357\,{\tiny[.32,.39]} & .291\,{\tiny[.25,.33]} & .285\,{\tiny[.25,.32]} & .288\,{\tiny[.25,.32]} \\
 & Common   & 1403 & .641\,{\tiny[.61,.67]} & .488\,{\tiny[.46,.52]} & .554\,{\tiny[.53,.58]} & .581\,{\tiny[.55,.61]} & .458\,{\tiny[.43,.49]} & .513\,{\tiny[.48,.54]} \\
\bottomrule
\end{tabular}
\caption{Tool-access ablation across all codes.}
\label{tab:tool-ablation-all}
\end{table*}

\begin{table*}[t]
\centering
\scriptsize
\setlength{\tabcolsep}{4pt}
\begin{tabular}{ll r ccc ccc}
\toprule
& & & \multicolumn{3}{c}{\textbf{Full-tools}} & \multicolumn{3}{c}{\textbf{No-tools}} \\
\cmidrule(lr){4-6} \cmidrule(lr){7-9}
\textbf{Model} & \textbf{Band} & \textbf{Sup} & \textbf{Precision} & \textbf{Recall} & \textbf{F1} & \textbf{Precision} & \textbf{Recall} & \textbf{F1} \\
\midrule
\multirow{3}{*}{Opus 4.6}
 & Rare     & 193 & .283\,{\tiny[.22,.34]} & .269\,{\tiny[.20,.34]} & .276\,{\tiny[.22,.33]} & .127\,{\tiny[.09,.17]} & .161\,{\tiny[.11,.21]} & .142\,{\tiny[.10,.19]} \\
 & Uncommon & 142 & .488\,{\tiny[.39,.60]} & .289\,{\tiny[.21,.37]} & .363\,{\tiny[.28,.45]} & .388\,{\tiny[.27,.51]} & .183\,{\tiny[.12,.25]} & .249\,{\tiny[.17,.33]} \\
 & Common   & 145 & .333\,{\tiny[.25,.42]} & .241\,{\tiny[.18,.31]} & .280\,{\tiny[.21,.35]} & .267\,{\tiny[.06,.50]} & .028\,{\tiny[.01,.06]} & .050\,{\tiny[.01,.10]} \\
\midrule
\multirow{3}{*}{Qwen 3.5 Plus}
 & Rare     & 193 & .234\,{\tiny[.16,.32]} & .207\,{\tiny[.15,.27]} & .220\,{\tiny[.16,.29]} & .102\,{\tiny[.06,.14]} & .119\,{\tiny[.07,.17]} & .110\,{\tiny[.06,.15]} \\
 & Uncommon & 142 & .437\,{\tiny[.31,.57]} & .218\,{\tiny[.15,.30]} & .291\,{\tiny[.20,.38]} & .279\,{\tiny[.15,.42]} & .085\,{\tiny[.04,.13]} & .130\,{\tiny[.07,.20]} \\
 & Common   & 145 & .286\,{\tiny[.20,.37]} & .221\,{\tiny[.15,.29]} & .249\,{\tiny[.18,.32]} & .214\,{\tiny[.00,.46]} & .021\,{\tiny[.00,.05]} & .038\,{\tiny[.00,.08]} \\
\midrule
\multirow{3}{*}{GLM-5 Turbo}
 & Rare     & 193 & .164\,{\tiny[.11,.22]} & .145\,{\tiny[.10,.19]} & .154\,{\tiny[.11,.20]} & .077\,{\tiny[.04,.13]} & .062\,{\tiny[.03,.10]} & .069\,{\tiny[.03,.11]} \\
 & Uncommon & 142 & .537\,{\tiny[.41,.68]} & .204\,{\tiny[.14,.27]} & .296\,{\tiny[.21,.38]} & .342\,{\tiny[.19,.50]} & .092\,{\tiny[.05,.15]} & .144\,{\tiny[.07,.22]} \\
 & Common   & 145 & .250\,{\tiny[.15,.35]} & .145\,{\tiny[.09,.21]} & .183\,{\tiny[.11,.25]} & .250\,{\tiny[.06,.50]} & .028\,{\tiny[.01,.06]} & .050\,{\tiny[.01,.10]} \\
\bottomrule
\end{tabular}
\caption{Tool-access ablation across multi-step codes.}
\label{tab:tool-ablation-multistep}
\end{table*}

% ===============================================================
\section{Resource Usage}
\label{app:resources}
% ===============================================================
Mean wall-clock latency (Lat.), estimated tokens per session in thousands (Tok.; characters$/4$), tool calls (Calls), voluntary submission rate (Sub.), and failure rate (Fail.). Failures combine timeouts, API errors, and stuck detection. No session reached the 200-turn or 200-tool-call limit.

\begin{table}[H]
\centering
\scriptsize
\setlength{\tabcolsep}{6pt}
\begin{tabular}{l r r r r r}
\toprule
\textbf{Model} & \textbf{Lat.} & \textbf{Tok.} & \textbf{Calls} & \textbf{Sub.} & \textbf{Fail} \\
& \textbf{(s)} & \textbf{(K)} & & \textbf{(\%)} & \textbf{(\%)} \\
\midrule
\multicolumn{6}{l}{\textit{No PLM-ICD tool}} \\
\addlinespace[1pt]
Claude Opus 4.6  & 204 & 25.6 & 64 & 100 &  0 \\
GPT-5.2          &  91 & 19.4 & 59 & 100 &  0 \\
GLM-5 Turbo      & 224 & 26.3 & 78 &  88 & 12 \\
Qwen 3.5 Plus    & 363 & 26.2 & 71 &  92 &  8 \\
Kimi K2.5        & 358 & 35.4 & 87 &  82 & 18 \\
Mistral Large 3   & 265 & 27.4 & 73 &  94 &  6 \\
\midrule
\multicolumn{6}{l}{\textit{With PLM-ICD tool}} \\
\addlinespace[1pt]
Opus 4.6 + PLM   & 176 & 24.5 & 57 & 100 &  0 \\
GPT-5.2 + PLM    &  65 & 23.9 & 79 & 100 &  0 \\
\bottomrule
\end{tabular}
\caption{Per-model resource usage.}
\label{tab:resource-usage}
\end{table}

% ===============================================================
\section{Discovery-Position Statistical Test}
\label{app:temporal-stats}
% ===============================================================
Statistical comparison of true-positive (TP) and false-positive (FP)
discovery positions across models. The table reports TP/FP sample counts ($n$), the median position difference ($\Delta$ Med.; FP minus TP), Cliff's $\delta$, and one-sided Mann--Whitney $U$ $p$ values
($H_1$: TP $<$ FP). Positive $\Delta$ Med. indicates that TP codes are
discovered earlier. Effect sizes follow \citet{Romano2006}.

\begin{table}[H]
\centering
\scriptsize
\setlength{\tabcolsep}{6pt}
\resizebox{\columnwidth}{!}{%
\begin{tabular}{l r r r r}
\toprule
\textbf{Model} & \textbf{$n$ (TP/FP)} & \textbf{$\Delta$ Med.} & \textbf{$\delta$} & \textbf{$p$} \\
\midrule
Claude Opus 4.6 & 1{,}026 / 770   & +0.079 & 0.233 & $<\!10^{-16}$ \\
GPT-5.2         &   972 / 955     & +0.106 & 0.200 & $<\!10^{-13}$ \\
GLM-5 Turbo     &   991 / 1{,}007 & +0.062 & 0.239 & $<\!10^{-19}$ \\
Mistral Large 3 &   904 / 1{,}544 & +0.163 & 0.261 & $<\!10^{-26}$ \\
Qwen 3.5 Plus   &   971 / 981     & +0.140 & 0.216 & $<\!10^{-15}$ \\
Kimi K2.5       &   993 / 1{,}753 & +0.095 & 0.301 & $<\!10^{-38}$ \\
\bottomrule
\end{tabular}}
\caption{Statistical test of the TP--FP discovery-position gap.}
\label{tab:temporal-position}
\end{table}

% ===============================================================
\section{Case Study: End-to-End Agentic Coding}
\label{app:casestudy}
% ===============================================================

Sample 11087585-DS-10 was coded by Claude Opus 4.6. The note describes a patient with right lower quadrant pain. CT showed a 7mm appendix with fat stranding and concurrent inflammation of the terminal ileum, cecum, and ascending colon. Discharge diagnosis:
\textit{colitis vs.\ acute uncomplicated appendicitis}. Past medical
history: OSA. Ground truth: \{K35.80, K52.9\}. The agent's trajectory on this sample is summarised below. The framework does not prescribe a fixed order; the steps reflect the agent's actual reasoning path.

\paragraph{Step 1: Index navigation and tabular verification.}
The agent searches the Alphabetical Index for \textit{Colitis},
\textit{Appendicitis $>$ acute}, and \textit{Apnea $>$ sleep $>$ obstructive}.
Browsing the K35 hierarchy confirms K35.80 over K35.2/K35.3, as no
peritonitis is documented.

\paragraph{Step 2: Guideline consultation.}
The agent retrieves Section~III.C (Uncertain Diagnosis), which instructs
that conditions connected by \textit{vs} should be coded as confirmed
for inpatient encounters. Both conditions are assigned.

\paragraph{Step 3: Exclusion decisions.}
R59.0 (reactive lymph nodes) and D72.829 (elevated WBC) are excluded as
findings integral to the inflammatory process per Guideline~I.B.5.
Z87.891 (nicotine history) is excluded as undocumented. G47.33 (OSA)
is included as a chronic PMH condition.

\begin{table}[H]
\centering
\scriptsize
\setlength{\tabcolsep}{4pt}
\begin{tabularx}{\columnwidth}{llX}
\toprule
\textbf{Code} & \textbf{Decision} & \textbf{Rationale} \\
\midrule
K52.9   & Include & Discharge dx; Section~III.C applies \\
K35.80  & Include & Discharge dx; CT confirms; no peritonitis \\
G47.33  & Include & Chronic PMH condition; inpatient convention \\
R59.0   & Exclude & Integral finding; Guideline~I.B.5 \\
D72.829 & Exclude & Lab finding; integral to process \\
Z87.891 & Exclude & Not documented in note \\
\bottomrule
\end{tabularx}
\caption{Inclusion and exclusion decisions for sample 11087585-DS-10.}
\end{table}

\paragraph{Outcome.}
Submitted: \{K52.9, K35.80, G47.33\}. Precision = 0.667, Recall =
1.000, F1 = 0.800. Both ground-truth codes correctly identified in
14 tool calls across 10 turns in 80 seconds. The false positive
G47.33 reflects genuine ambiguity: the agent's reasoning is defensible
under inpatient coding conventions, but annotators excluded PMH-only
conditions not actively managed during the encounter.

% ===============================================================
\section{System Prompt}
\label{app:prompt}
% ===============================================================

The following is the complete system prompt used for all agentic
evaluation runs. The prompt is rendered at runtime with a single
boolean variable controlling whether the PLM-ICD priming step is
included. For no-PLM configurations, Steps~1--2 and the
\texttt{suggest\_candidate\_codes} tool reference are omitted;
all coding rules remain identical.

\scriptsize
\begin{verbatim}
You are an expert ICD-10-CM medical coder. Read a clinical
note and assign ALL applicable ICD-10-CM codes using ONLY
the reference tools provided. You must NOT rely on memorized
codes — every code you submit must be discovered through
the tools.

## Coding Workflow

### Step 1: Prime with PLM-ICD and read the note
Call suggest_candidate_codes() first. These candidates are
your primary coding checklist — verify each one systematically.
Then read the note to identify conditions PLM-ICD may have
missed (Z-codes, external causes, medications).

### Step 2: Verify every PLM-ICD candidate
For each candidate, you MUST:
1. Navigate to the specific code via the Alphabetical Index
2. Drill to the most specific billable code with
   browse_code_children
3. Verify with verify_code_in_tabular
4. Submit unless the note clearly contradicts the code

Apply Guideline I.B.18: use the unspecified code (.9) unless
the note explicitly documents the specific subtype.

### Step 3: Discover additional codes
Re-read the note section by section:
- Discharge diagnoses: every listed diagnosis
- Medications: code the underlying condition ONLY if actively
  administered/managed during the encounter
- Past medical history: ONLY chronic conditions actively
  managed or evaluated during this encounter
- Vitals: BMI if documented (Z68.x)
- Family history: if it influenced care (Z80-Z84)
- Social history: tobacco use (F17/Z87.891), alcohol if relevant
- Procedures: T-codes or injury codes needing external cause
  codes (Y83-Y84)?

Also check for companion codes ("Use additional code",
"Code first", "Code also") — these are mandatory.

MANDATORY external cause sequence — if you have ANY T-code
(T80-T88), S-code, or W/V/X/Y code, assign:
1. Cause/event: browse Y83 for surgical complications, or
   search_external_cause_index for injuries
2. Place of occurrence (Y92.-): assign Y92.9 if not stated
3. Match 7th character to the injury/T-code's 7th character

### Step 4: Navigate and verify
Search: search_alphabetical_index(term) with the condition
noun. Use > syntax to drill sub-terms.
Special tables: search_neoplasm_table, search_drug_table,
search_external_cause_index.
Drill: browse_code_children(code) to reach a leaf code.
  - Laterality: match documented side exactly
  - Specificity: use unspecified if not further qualified
  - When in doubt between siblings, prefer less specific
    (Guideline I.B.18)
Verify: verify_code_in_tabular on every leaf code.

### Step 5: Submit
Before submitting, apply these filters:
- Section III: only code conditions that affected patient
  care during THIS encounter
- Do NOT code symptoms integral to a confirmed diagnosis
- Do NOT code incidental imaging findings not acted upon
- Hypertension + CKD/heart failure: use I11.x/I12.x/I13.x,
  not I10 separately
- Diabetes with complications: use combination codes
  (E11.22, E11.40, etc.), not E11.9 + separate complication

## Key Coding Rules

Specificity (Guideline I.B.18): code to the level of
certainty known. Use .9 (unspecified) when the record does
not support a more specific code. Use .8 (other specified)
only if the note explicitly describes a subtype mapping to .8.

Code structure: codes are 3-7 characters. Use placeholder X
for empty positions before the 7th character (e.g.,
T81.44XA, W01.0XXA).

Etiology/manifestation:
- "Code first" = sequence the underlying condition first
- "Use additional code" = add a secondary code after
- "Code also" = two codes may be required
- Manifestation codes in brackets [ ] in the Index: always
  submit BOTH codes

"With"/"And": "with" or "in" in the Index or Tabular = causal
relationship presumed, code as related without explicit
provider documentation. "And" = "and/or".

External cause codes (V00-Y99): assign cause/event, place
of occurrence (Y92.-), and activity (Y93.-) if documented.

Uncertain diagnoses:
- Inpatient: "probable," "suspected," "likely," "rule out"
  -> code as if confirmed
- Outpatient: do NOT code uncertain diagnoses

## Tool Reference

suggest_candidate_codes(top_k, threshold)
  PLM-ICD neural code suggestions. Call this first.

search_alphabetical_index(term, page)
  ICD-10-CM Alphabetical Index. Use > for drilling
  (e.g., Failure > heart > congestive).

search_neoplasm_table(site)
  Search by anatomical site for neoplasm codes.

search_drug_table(substance)
  Search by substance for drug-related codes.

search_external_cause_index(cause)
  Search for external causes, places, and activities.

browse_code_children(code)
  List child codes. Use on every code from the Index.

verify_code_in_tabular(code)
  Full Tabular List entry. Always verify before submitting.

lookup_coding_guidelines(query, section, page)
  Official Coding Guidelines by section.

submit_codes(codes, sequencing_notes)
  Submit final answer.

## Your Task
Read the clinical note. Call suggest_candidate_codes(),
verify each candidate, re-read the note for missed
conditions, verify all codes in the Tabular List, then
submit. Capture every codeable condition.
\end{verbatim}
\scriptsize

\end{document}